\documentclass[letterpaper, 10 pt, conference]{ieeeconf}
\IEEEoverridecommandlockouts    
\usepackage{graphics}           
\usepackage{times}              
\usepackage{amsmath}            
\usepackage{amssymb}            
\usepackage{graphicx}
\usepackage{algorithm}
\usepackage[noend]{algpseudocode}
\usepackage{booktabs}
\usepackage{color}
\definecolor{instructioncolor}{rgb}{.5,.5,.5}

\usepackage[font=small]{caption}

\def\secref#1{Sec.~\ref{#1}}
\def\figref#1{Fig.~\ref{#1}}

\def\eqref#1{Eq.~(\ref{#1})}

\makeatletter
\usepackage{xspace}
\DeclareRobustCommand\onedot{\futurelet\@let@token\@onedot}
\def\@onedot{\ifx\@let@token.\else.\null\fi\xspace}
 
\def\ie{i.e\onedot}

\def\etal{{et al}\onedot}
\makeatother

\def\etalcite#1{\etal~\cite{#1}}

\usepackage{array}
\newcolumntype{L}[1]{>{\raggedright\let\newline\\\arraybackslash\hspace{0pt}}m{#1}}
\newcolumntype{C}[1]{>{\centering\let\newline\\\arraybackslash\hspace{0pt}}m{#1}}
\newcolumntype{R}[1]{>{\raggedleft\let\newline\\\arraybackslash\hspace{0pt}}m{#1}}

\renewcommand{\b}[1]{\mbox{\boldmath$#1$}}

\renewcommand{\v}[1]{{\b #1}} 

\usepackage{color}
\usepackage{multicol}
\definecolor{CommentPink}{rgb}{1,0.2,0.5}
\definecolor{CommentBlue}{rgb}{0,0,1}
\definecolor{CommentGreen}{rgb}{0,1,0}

\title{\LARGE \bf Adaptive Path Planning for UAV-based\\Multi-Resolution Semantic Segmentation}

\author{Felix Stache$^*$ \and Jonas Westheider$^*$ \and Federico Magistri \and Marija Popovi\'{c} \and Cyrill Stachniss
  \thanks{$^*$: authors with equal contribution.}
  \thanks{All authors are with the University of Bonn, Germany.}%
  \thanks{This work has been funded by the Deutsche Forschungsgemeinschaft (DFG, German Research Foundation) under Germany's Excellence Strategy, EXC-2070 -- 390732324 (PhenoRob).}
}

\begin{document}
\maketitle
\thispagestyle{empty}
\pagestyle{empty}

\begin{abstract}
 %

In this paper, we address the problem of adaptive path planning for accurate semantic segmentation of terrain using unmanned aerial vehicles (UAVs).
The usage of UAVs for terrain monitoring and remote sensing is rapidly gaining momentum due to their high mobility, low cost, and flexible deployment. However, a key challenge is planning missions to maximize the value of acquired data in large environments given flight time limitations.
To address this, we propose an online planning algorithm which adapts the UAV paths to obtain high-resolution semantic segmentations necessary in areas on the terrain with fine details as they are detected in incoming images. This enables us to perform close inspections at low altitudes only where required, without wasting energy on exhaustive mapping at maximum resolution. A key feature of our approach is a new accuracy model for deep learning-based architectures that captures the relationship between UAV altitude and semantic segmentation accuracy. We evaluate our approach on the application of crop/weed segmentation in precision agriculture using real-world field data.

\end{abstract}

\section{Introduction}
\label{sec:intro}

Unmanned aerial vehicles (UAVs) are experiencing a rapid uptake in a variety of aerial monitoring applications, including search and rescue~\cite{Meera19}, wildlife conservation~\cite{Manfreda2018}, and precision agriculture~\cite{popovic2017icra,popovic2017iros,Vivaldini2019}. They offer a flexible and easy to execute a way to monitor areas from a top-down perspective. Recently, the advent of deep learning has unlocked their potential for image-based remote sensing, enabling flexible, low-cost data collection and processing~\cite{Carrio2017}. However, a key challenge is planning paths to efficiently gather the most useful data in large environments, while accounting for the constraints of physical platforms, e.g. on fuel/energy, as well as the on-board sensor properties.

This paper examines the problem of deep learning-based semantic segmentation using UAVs and the exploitation of this information in path planning. Our goal is to adaptively select the next sensing locations above a 2D terrain to maximize the classification accuracy of objects or areas of interest seen in images, e.g. animals on grassland or crops on a field. This enables us to perform targeted high-resolution classification only where necessary and thus maximize the value of data gathered during a mission.

Most data acquisition campaigns rely on coverage-based planning to generate UAV paths at a fixed flight altitude~\cite{Cabreira2019}. Although easily implemented, the main drawback of such methods is that they assume an even distribution of features in the target environment; mapping the entire area at a constant image spatial resolution governed by the altitude. Recent work has explored \textit{informative planning} for terrain mapping, whereby the aim is to maximize an information-theoretic mapping objective subject to platform constraints. However, these studies either consider 2D planning at a fixed altitude or apply simple heuristic predictive sensor models~\cite{popovic2017icra,popovic2017iros,Meera19}, which limits the applicability of future plans. A key challenge is reliably characterizing how the accuracy of segmented images varies with the altitude and relative scales of the objects in registered images.

\begin{figure}[t]
 \centering
 \includegraphics[width=0.48\linewidth]{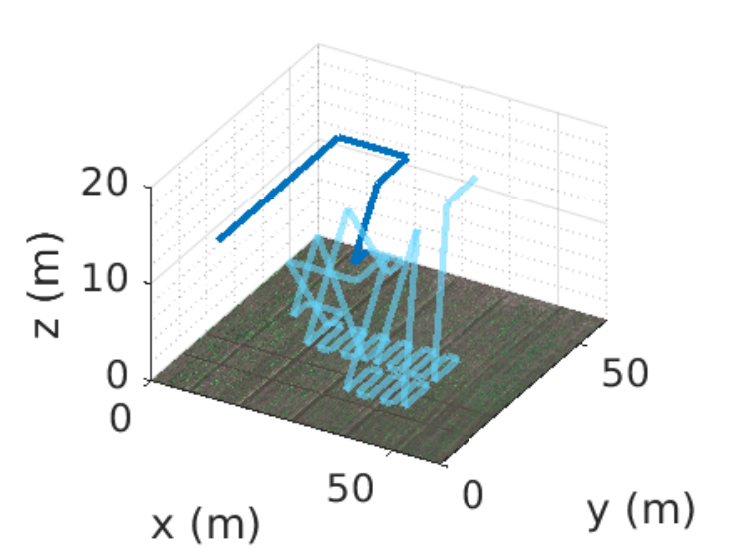}
 \includegraphics[width=0.48\linewidth]{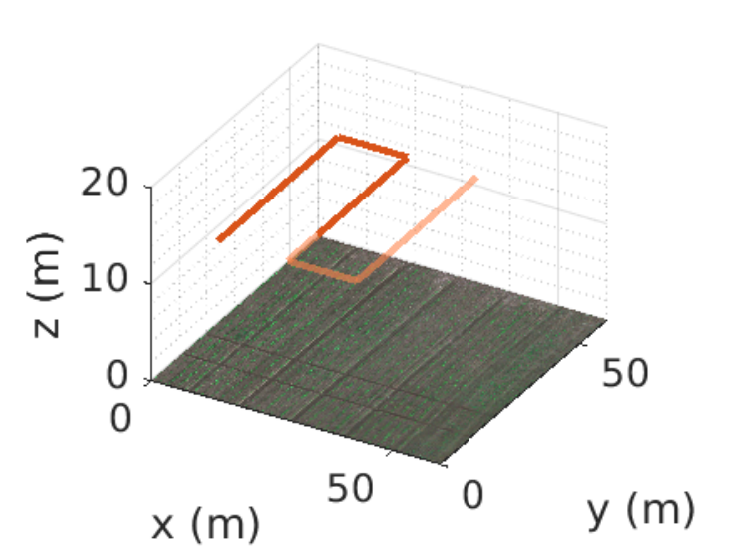}
 \vspace{1mm}

 \includegraphics[width=0.95\linewidth]{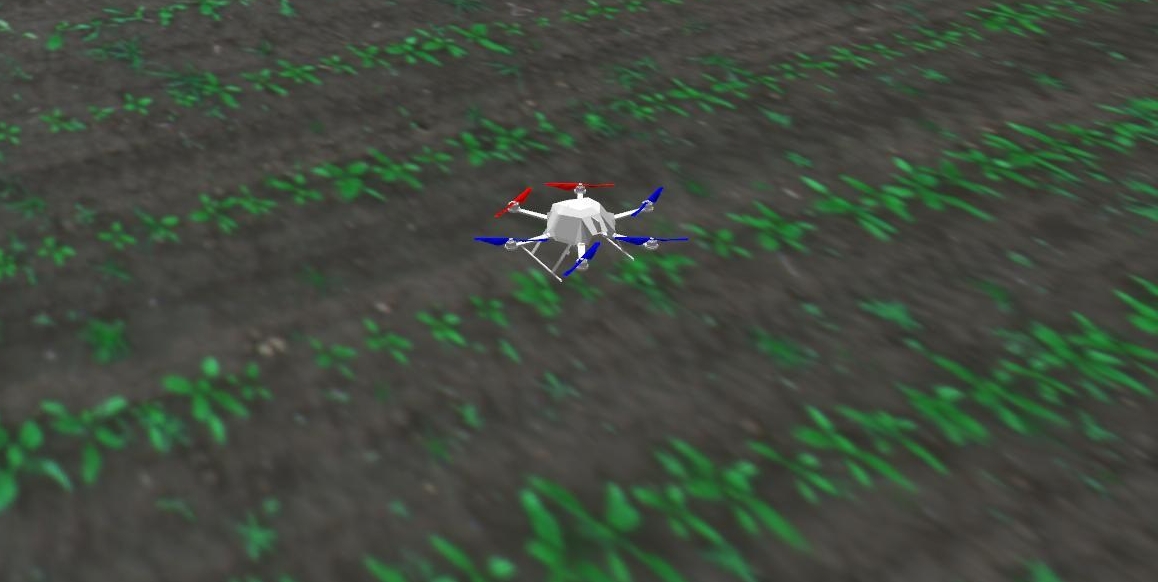}
 \caption{A comparison of our proposed adaptive path planning strategy (top-left) against lawn-mower coverage planning (top-right) for UAV-based field segmentation, evaluated on the application of precision agriculture using real field data (bottom). By allowing the paths to change online, our approach enables selecting high-resolution (low-altitude) imagery in areas with more semantic detail, enabling higher-accuracy, fine-grained segmentation in these regions.}
 \label{fig:motivation}
\end{figure}

To address this, we propose a new adaptive planning algorithm that directly tackles the altitude dependency of the deep learning semantic segmentation model using UAV-based imagery. First, our approach leverages prior labeled terrain data to empirically determine how classification accuracy varies with altitude; we train a deep neural network with images obtained at different altitudes that we use to initialize our planning strategy. 
Based on this analysis, we develop a \textit{decision function} using Gaussian Process (GP) regression that is first initialized on a training field and then updated online on a separate testing field during a mission as new images are received. For replanning, the UAV path is chosen according to the decision function and segmented images to obtain higher classification accuracy in more semantically detailed or interesting areas. This allows us to gather more accurate data in targeted areas without relying on a heuristic sensor model for informative planning.

The contributions of this work are: (i) an online planning algorithm for UAVs that uses the semantic content of new images to adaptively map areas of finer detail with higher accuracy. (ii) A variable-altitude accuracy model for deep learning-based semantic segmentation architectures and its integration in our planning algorithm. (iii) The evaluation of our approach against state-of-the-art methods using real-world data from an agricultural field to demonstrate its performance.
We note that, while this work targets the application of precision agriculture, our algorithm can be used in any other UAV-based semantic segmentation scenario, e.g., search and rescue~\cite{Meera19}, urban scene analysis, wetland assessment, etc.

\section{Related Work}
\label{sec:related}


There is an emerging body of literature addressing mission planning for UAV-based remote sensing. This section briefly reviews the sub-topics most related to our work.

\textbf{UAV-based Semantic Segmentation:} The goal of semantic segmentation is to assign a predetermined class label to each pixel of an image. State-of-the-art approaches are predominantly based on convolutional neural networks (CNNs) and have been successfully applied to aerial datasets in various scenarios~\cite{Sa2018,Carrio2017,Nguyen2019,Lyu2020,Sa2018b}. In the past few years, technological advancements have enabled efficient segmentation on board small UAVs with limited computing power. Nguyen~\etalcite{Nguyen2019} introduced MAVNet, a light-weight network designed for real-time aerial surveillance and inspection. Sa~\etalcite{Sa2018} and Deng~\etalcite{Deng2020} proposed CNN methods to segment vegetation for smart farming using similar platforms. Our work shares the motivation of these studies; we adopt ERFNet~\cite{romera17} to perform efficient aerial crop/weed classification in agricultural fields. However, rather than flying predetermined paths for monitoring, as in previous studies, we focus on planning: we aim to exploit modern data processing capabilities to localize areas of interest and finer detail (e.g. high vegetation cover) online and steer the robot for adaptive, high-accuracy mapping in these regions.

\textbf{Adaptive Path Planning:} Adaptive algorithms for active sensing allow an agent to replan online as measurements are collected during a mission to focus on application-specific interests. Several works have successfully incorporated adaptivity requirements within informative path planning problems. Here, the objective is to minimize uncertainty in target areas as quickly as possible, e.g. for exploration~\cite{stachniss2005rss}, underwater surface inspection~\cite{Hollinger2013}, target search~\cite{Meera19,Sadat15,Singh2009}, and environmental sensing~\cite{Singh2009}. These problem setups differ from ours in several ways. First, they consider a probabilistic map to represent the entire environment, using a sensor model to update the map with new uncertain measurements. In contrast, our approach directly exploits the accuracy in semantic segmentation to drive adaptive planning. Second, they consider a predefined, i.e. non-adaptive, sensor model, whereas ours is adapted online according to the behavior of the semantic segmentation model.

Very few works have considered planning based on semantic information. Bartolomei~\etalcite{Bartolomei20} introduced a perception-aware planner for UAV pose tracking. Although like us, they exploit semantics to guide next UAV actions, their goal is to triangulate high-quality landmarks whereas we aim to obtain accurate semantic segmentation in dense images. Dang~\etalcite{Dang2018} and Meera~\etalcite{Meera19} study informative planning for target search using object detection networks. Most similar to our approach is that of Popovi\'{c}~\etalcite{Popovic2020AURO}, which adaptively plans the 3D path of a UAV for terrain monitoring based on an empirical performance analysis of a SegNet-based architecture at different altitudes~\cite{Sa2018}. A key difference is that our decision function, representing the network accuracy, is not static. Instead, we allow it to change online and thus adapt to new unseen environments.

\begin{figure}[t]
  \centering
  \includegraphics[width=0.99\linewidth]{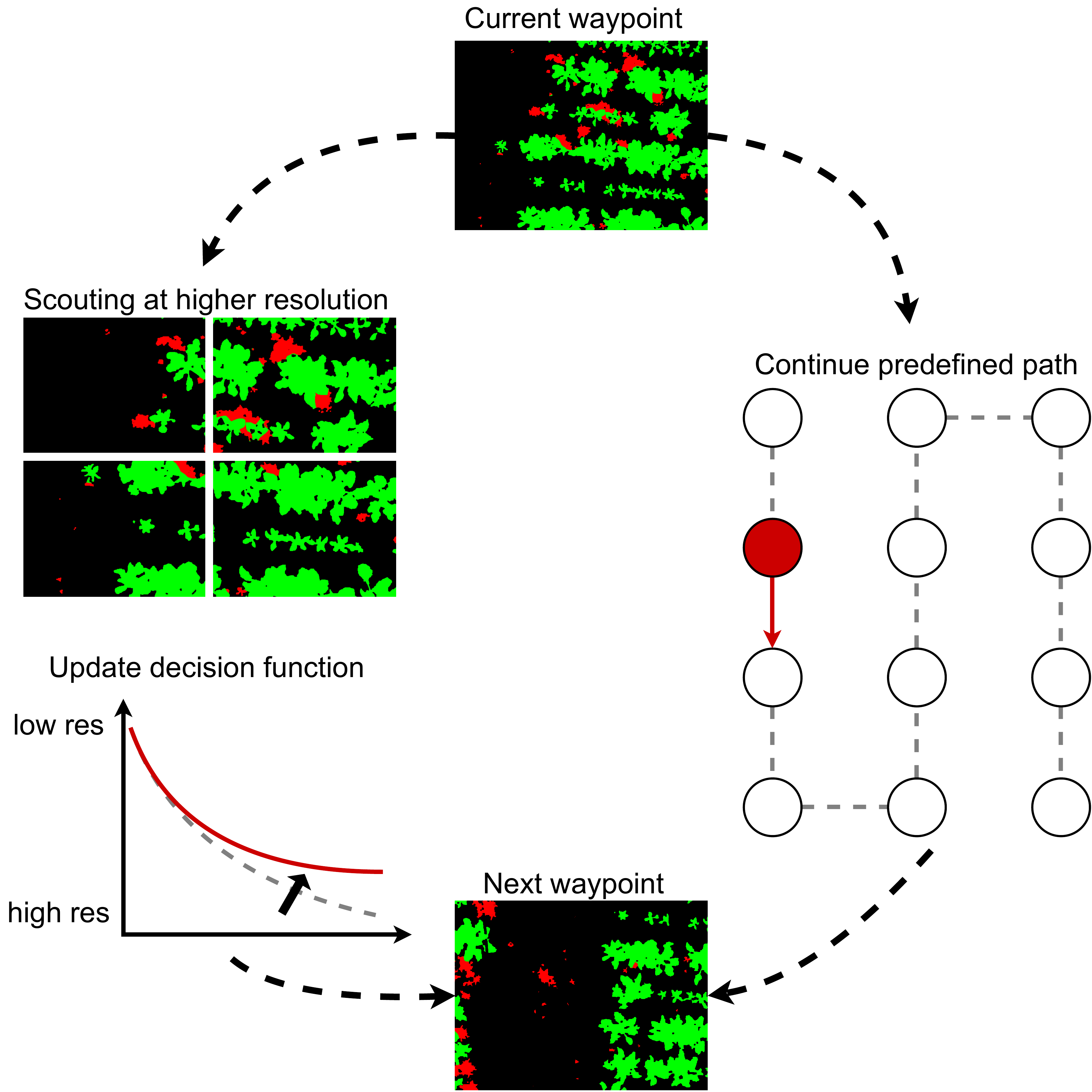}
  \caption{Each time the UAV segments one region of the field, we decide if the UAV should follow its predefined path (right side) or if it should scout the same region with a lower altitude, i.e. obtaining images with high resolution (left side). In the second case, we update the decision strategy by comparing the segmentation results of the same regions at different altitudes.}
  \label{fig:workflow}
\end{figure}

\textbf{Multi-Resolution:} An important trade-off in aerial imaging arises from the fact that spatial resolution degrades with increasing coverage, e.g. Pe{\~{n}}a~\etalcite{Pena2015} show that there are optimal altitudes for monitoring plants based on their size. Relatively limited research has tackled this challenge in the contexts of semantic segmentation and planning. For mapping, Dang~\etalcite{Dang2018} employ an interesting method for weighting distance measurements according to their resolution. Sadat~\etalcite{Sadat15} propose an adaptive coverage-based strategy that assumes sensor accuracy increases with altitude. Other studies~\cite{Vivaldini2019,Sa2018b} only consider fixed-altitude mission planning. We follow previous approaches that empirically assess the effects of multi-resolution observations for trained models~\cite{Meera19,Popovic2020AURO,Qingqing2020}. Specifically, our contribution is a new decision function that supports online updates for more reliable predictive planning. 


\section{Our Approach}
\label{sec:main}

The goal of this work is to maximize the accuracy in the semantic segmentation of RGB images taken by a camera on-board a UAV with a limited flight time. We propose a data-driven approach that uses information from incoming images to adapt an initial predefined UAV flight path online. The main idea behind our approach is to guide the UAV to take high-resolution images for fine-grained segmentation at lower altitudes (higher resolutions) only where necessary.

As a motivating application, our problem setup considers a UAV monitoring an agricultural field to identify crops and weeds for precision treatment.
We first divide the target field into non-overlapping regions and, for each, associate a waypoint in the 3D space above the field from which the camera footprint of the UAV camera covers the entire area. From these waypoints, we then define a lawn-mower coverage path that we use to bootstrap the adaptive strategy. Our strategy consists of two steps. First, at each waypoint along the lawnmower path, we use a deep neural network to assign a semantic label to each pixel in the observed region (soil, crop, and weed in our selected use-case). Second, based on the segmented output, we decide whether the current region requires more detailed re-observation at a higher image resolution, \ie, lower UAV altitude; otherwise, the UAV continues its pre-determined coverage path. 

A key aspect of our approach is a new data-driven decision function that enables the UAV to select a new altitude for higher-resolution images if they are needed.
This decision function is updated adaptively during the mission by comparing the segmentation results of the current region at the different altitudes. This enables us to precisely capture the relationship between image resolution (altitude) and segmentation accuracy when planning new paths.
\figref{fig:workflow} shows an overview of our planning strategy. In the following sub-sections, we describe the CNN for semantic segmentation and the path planning strategy, which consists of offline planning and online path adaptation.


\subsection{Semantic Segmentation}
\label{subsec:segmentation}
In this work, we consider the semantic segmentation of RGB images not only as of the final goal but also as a tool to define adaptive paths for re-observing given regions of the field.
Each time the UAV reaches a waypoint, we perform a pixel-wise semantic segmentation to assign a label (crop, weed or soil in our case) to each pixel in the current view. We use the ERFNet~\cite{romera17} architecture provided by the Bonnetal framework~\cite{milioto2019ieee} that allows for real-time inference. We train this neural network on RGB images collected at different altitudes to allow it to generalize across possible altitudes without the need for retraining. 
If the same region is observed by the camera from different altitudes, we preserve the results obtained with the highest resolution, assuming that higher-resolution images yield greater segmentation accuracy.

\subsection{Path Planning: Basic Strategy}
\label{subsec:planning}

The initial flight path is calculated based on the standard lawn-mower strategy~\cite{Galceran2013}. Such a path enables covering the region of interest efficiently without any prior knowledge. We aim to adapt this path according to the non-uniform distribution of features in the field to improve semantic segmentation performance.

For a desired region of interest, we define a lawn-mower path based on a series of waypoints. A waypoint is defined as a position $\v{w_i}$ in the 3D UAV workspace where: (i) the UAV camera footprint does not overlap the footprints of any other waypoint; (ii) the UAV performs the semantic segmentation of its current field of view; (iii) the UAV decides to revise its path or to execute the path as previously determined; and (iv) we impose zero velocity and zero acceleration.

The initial flight path is calculated in form of fixed waypoints at the highest altitude $ \v{W}^{h_\text{max}} = \{ \v{w_0}, \v{w_1},\ldots, \v{w_n} \} $. If necessary, we modify this coarse plan by inserting further waypoints based on the new imagery as it arrives. At each waypoint $\v{w_i}$, UAV decides either to follow the pre-determined path, \ie moving to $\v{w_{i+1}}$, or to inspect the current region more closely at a lower altitude. In the second case, we define a second series of waypoints, $ \v{W}^{h^\prime} = \{ \v{w_0}, \v{w_1}, \ldots, \v{w_n} \} $, at the desired altitude, $h^\prime$, that will be inserted before $\v{w_{i+1}} \in \v{W}^{h_\text{max}}$ so that the resulting path, at the desired altitude, is a lawn-mower strategy covering the camera footprint from $\v{w_{i}} \in \v{W}^{h_\text{max}}$.

\subsection{Planning Strategy: Offline Initialization}
\label{subsec:init}
We develop a decision function that takes a given waypoint as input and outputs the next waypoint, either $\v{w_{i+1}} \in \v{W}^{h_\text{max}}$ or $\v{w_{0}} \in \v{W}^{h^\prime}$, given the segmentation result. In the case of an altitude change, our decision function outputs also the value of the desired altitude $h^\prime$. To do this, we start by defining a vegetation ratio providing the number of pixels classified as vegetation (crop and weed) as a fraction of the total number of pixels in the image:

\begin{equation}\label{eq:vr} 
v = \frac{\sum_{c \in \{ \text{crop, weed} \} } p_c }{p_{\text{tot}}},
\end{equation}
where $p_{tot}$ is the total number of pixel and $p_c$ is the total number of pixels classified as $c$. This vegetation ratio gives us a way to infer how valuable it is to spend time on the current region of the field. It captures the intuition that higher values of this ratio indicate more possible misclassifications between the crop and weed classes. To quantify such a relationship, we let the UAV run on a separate field, where we have access to ground truth data, segmenting regions of the fields with different altitudes. Segmenting the same region of the field with different altitudes provides two pieces of information that we use to shape the decision function. On one hand, we have the difference between the altitudes from which we segment the field, $\Delta h = h_\text{max} - h^\prime$. On the other hand, we have the the difference between the vegetation ratio in the predicted segmentation, $\Delta v = v_{h_\text{max}} - v_{h^\prime}$. At the same time, we can compare the vegetation ratio to the accuracy of the predicted segmentation by computing the mean intersection over union (mIoU). Where the mIoU is defined as the average over the classes $C=\{ \text{crop, weed, soil} \}$ of the ratio between the intersection of ground truth and predicted segmentation and the union of the same quantities:

\begin{figure}[t]
  \centering
   \includegraphics[width=0.99\linewidth]{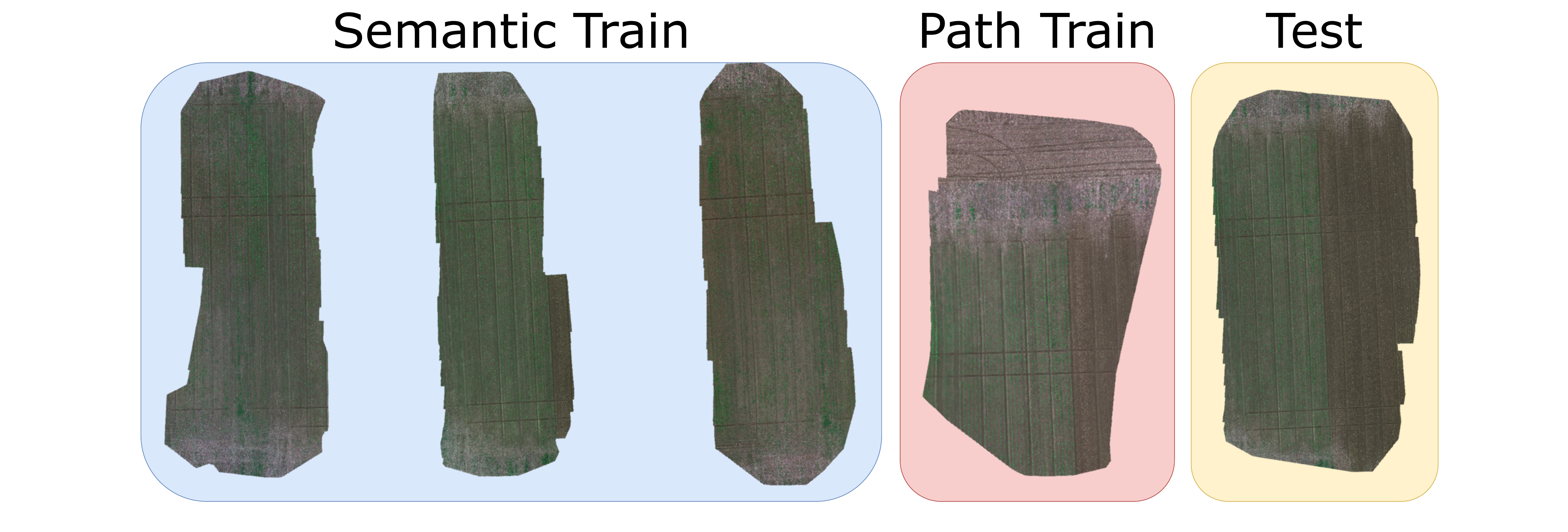}
  \caption{We use three different fields from the WeedMap dataset \cite{Sa2018b} to train a CNN for semantic segmentation and one field to initialize the planning strategy. The remaining field is used for evaluation. For an extensive evaluation of our approach, we swap the roles of the fields so that we test our algorithm on each field once.}
  \label{fig:dataset_split}  
\end{figure}

\begin{equation}\label{eq:iou}
\text{mIoU} = \frac{1}{|C|} \sum_{c \in C}  \frac{\text{gt}_\text{c} \cap \text{prediction}_\text{c}}{\text{gt}_\text{c} \cup \text{prediction}_\text{c}}.
\end{equation}
Again, we define the difference between mIoUs at different altitudes as, $\Delta \text{mIoU} = \text{mIoU}_{h_\text{max}} - \text{mIoU}_{h^\prime}$ .

Our method thus considers two sets of observations, representing the relationships between the vegetation ratio and UAV altitude ($\mathcal{O}$) and between vegetation ratio and mIoU ($\mathcal{I}$) as follows:
\begin{multicols}{2}
\begin{equation*}\label{eq:veg2h}
  \mathcal{O} = \begin{bmatrix}
   \Delta v_0 & \Delta h_0\\ 
   \Delta v_1 & \Delta h_1\\ 
   \multicolumn{2}{c}{\vdots}\\
   \Delta v_n & \Delta h_n\\ 
  \end{bmatrix},
\end{equation*}

\begin{equation*}\label{eq:veg2h}
  \mathcal{I} = \begin{bmatrix}
   \Delta v_0 & \Delta \text{mIoU}_0\\ 
   \Delta v_1 & \Delta \text{mIoU}_1\\ 
   \multicolumn{2}{c}{\vdots}\\
   \Delta v_n & \Delta \text{mIoU}_n\\ 
  \end{bmatrix}.
\end{equation*}

\end{multicols}

While both sets are initialized offline, we only update $\mathcal{O}$ online given that $\mathcal{I}$ requires access to ground truth that is clearly not available on testing fields. We fit both sets of observations using GP regression \cite{Rasmussen2006}. A GP for a function~$f(x)$ is defined by a mean function~$m(x)$ and a covariance function~$k(x_i,\,x_j)$:

\begin{equation}\label{eq:prior}
f(x) \thicksim GP(m(x),k(x_i,x_j)).
\end{equation}

A common choice is to set the mean function~$m(x) = 0$ and to use the squared exponential covariance function:
\begin{eqnarray}
\label{eq:sqexp}
  k(x_i,x_j) &=& \varsigma_{f}^2 \mathrm{exp}\left(-\frac{1}{2}\frac{|x_i-x_j|^2}{\ell^2}\right) +\varsigma_{n}^2,
\end{eqnarray}
\noindent where $\mathbf{\theta} =\{\ell,\,\varsigma_{f}^2,\,\varsigma_{n}^2\}$ are the model hyperparameters and represent respectively the length scale~$\ell$, the variance of the output~$\varsigma_{f}^2$ and of the noise~$\varsigma_{n}^2$.
Typically, the hyperparameters are learned from the training data by maximizing the log marginal likelihood.
Given a set of observations~$y$ of~$f$ for the inputs~$\v{X}$ (\ie our sets $\mathcal{O}$, $\mathcal{I}$), GP regression allows for learning a predictive model of~$f$ at the query inputs~$\v{X}_{*}$ by assuming a joint Gaussian distribution over the samples. 
The predictions at~$\v{X}_{*}$ are represented by the predictive mean~$\mu_{*}$ and variance~$\sigma_{*}^2$ defined as:

\begin{eqnarray}
  \begin{aligned}
    \mu_{*} =&~\v{K}(\v{X}_{*},\v{X})\,\v{K}_{\mathrm{XX}}^{-1}\,y, \\
    \sigma_{*}^2 =&~\v{K}(\v{X}_{*},\v{X}_{*})\,-\v{K}(\v{X}_{*},\v{X})\,{\v{K}_{\mathrm{XX}}}^{-1}\,\v{K}(\v{X},\v{X}_{*}),
  \end{aligned}
  \label{eq:gp}
\end{eqnarray}

\noindent where~$\v{K}_{\mathrm{XX}} = \v{K}(\v{X},\v{X})+\varsigma_{n}^2\,\v{I}$, and $\v{K}(\cdot,\,\cdot)$ are matrices constructed using the covariance function~$k(\cdot,\cdot)$ evaluated at the training and test inputs, $\v{X}$ and~$\v{X}_{*}$.
In the following, we will use the ground sampling distance (GSD) to identify the image resolution (thus the UAV altitude) from which the UAV performs the semantic segmentation. The GSD is defined as: $\text{GSD} = \frac{h \* S_w}{f \* I_w }$, where $h$ is the UAV altitude, $S_w$ the sensor width of the camera in mm, $f$ the focal length of the camera in mm and $I_w$ the image width in pixels. 


\subsection{Planning Strategy: Online Adaptation}
\label{subsec:online}
To adapt the UAV behavior online to fit the differences between the testing and training fields, we update the GP defined by the set $\mathcal{O}$ in the following way. In the testing field, each time the UAV decides to change altitude to a lower one, we compute a new pair $\Delta v' , \Delta h'$ and re-compute the GP output as defined in \eqref{eq:prior}.

\begin{figure}[t]
  \centering 
   \includegraphics[width=0.9\linewidth]{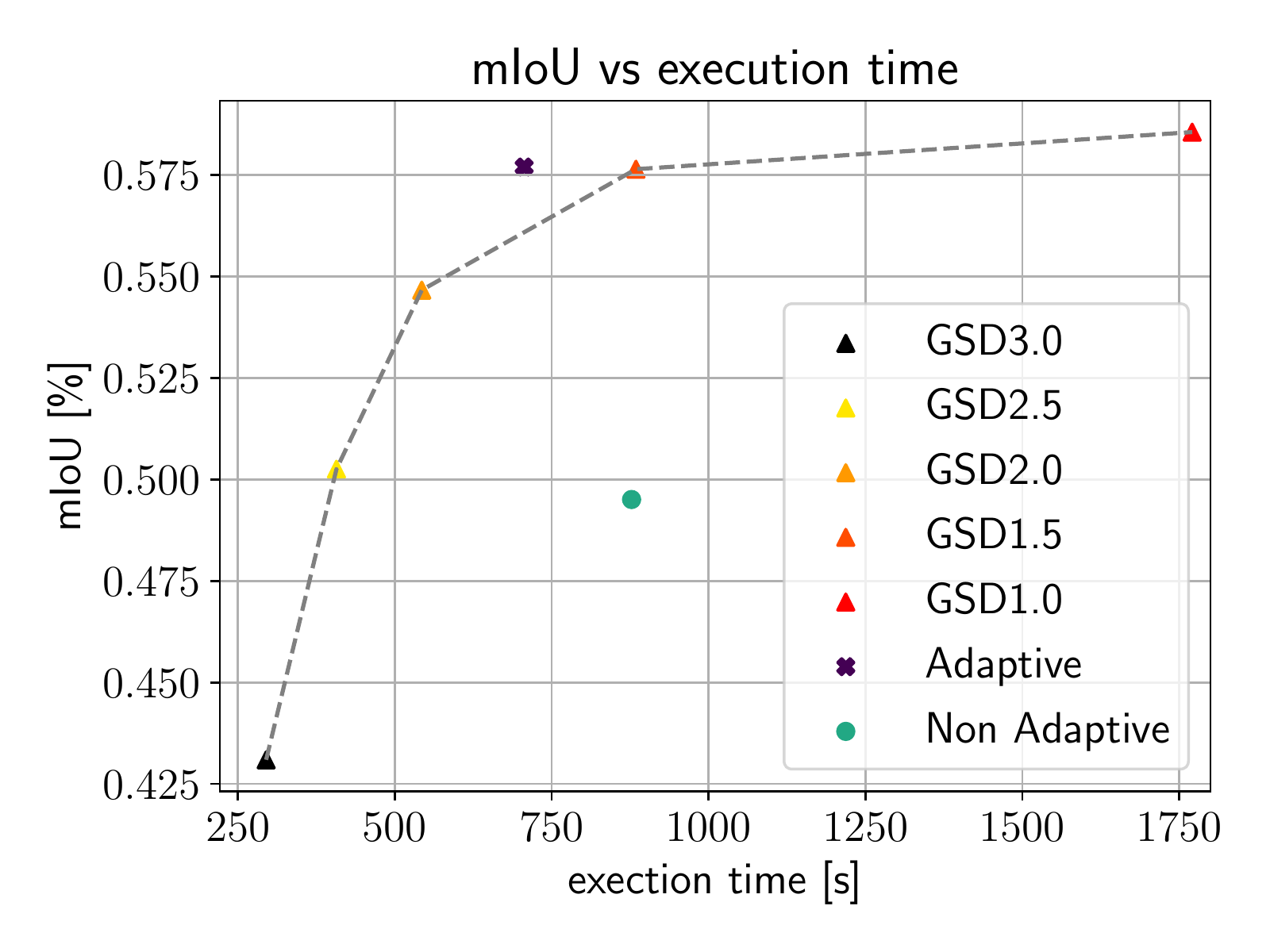}
  \caption{The averaged results for the testing fields from the WeedMap dataset. The blue square lies to the left of all performances with a linear decision function, indicating some performance improvement.}
  \label{fig:results}
\end{figure}

\begin{figure*}[t]
  \centering
   \includegraphics[width=0.3\linewidth]{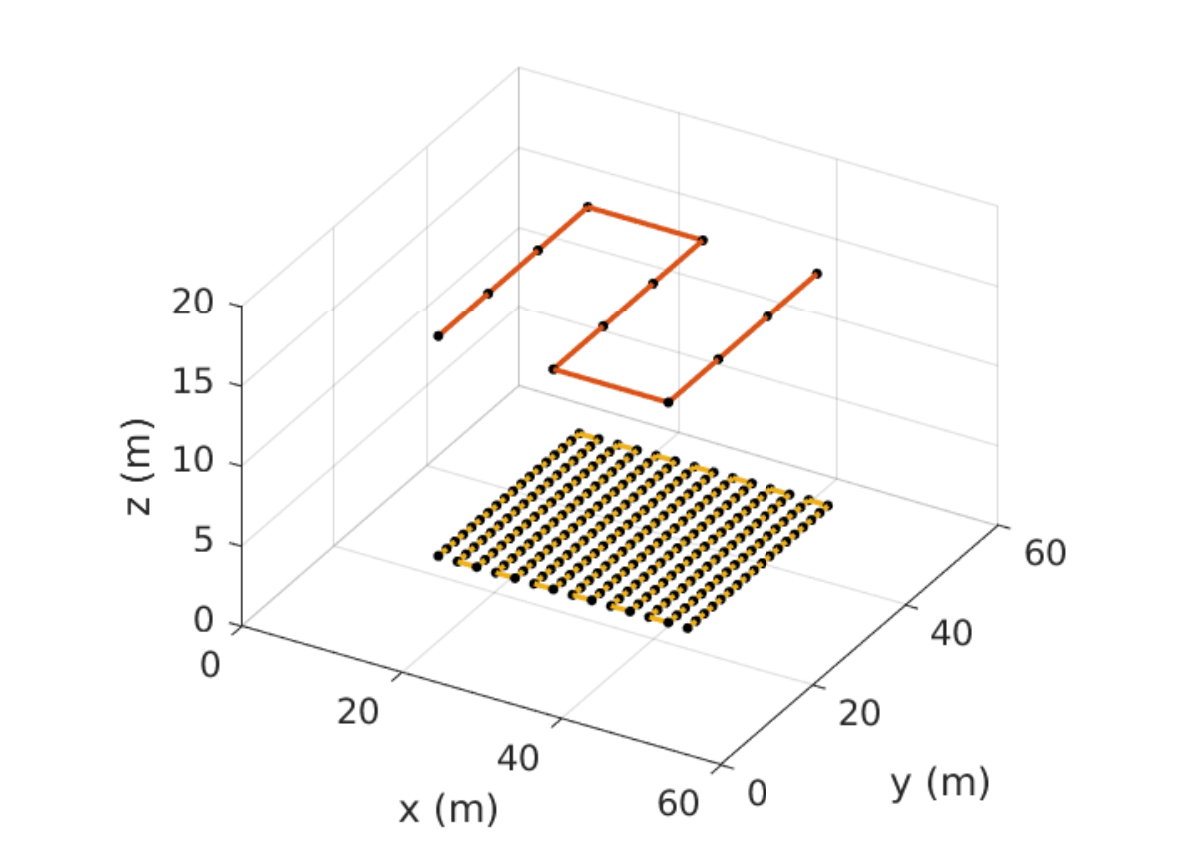}
   \includegraphics[width=0.3\linewidth]{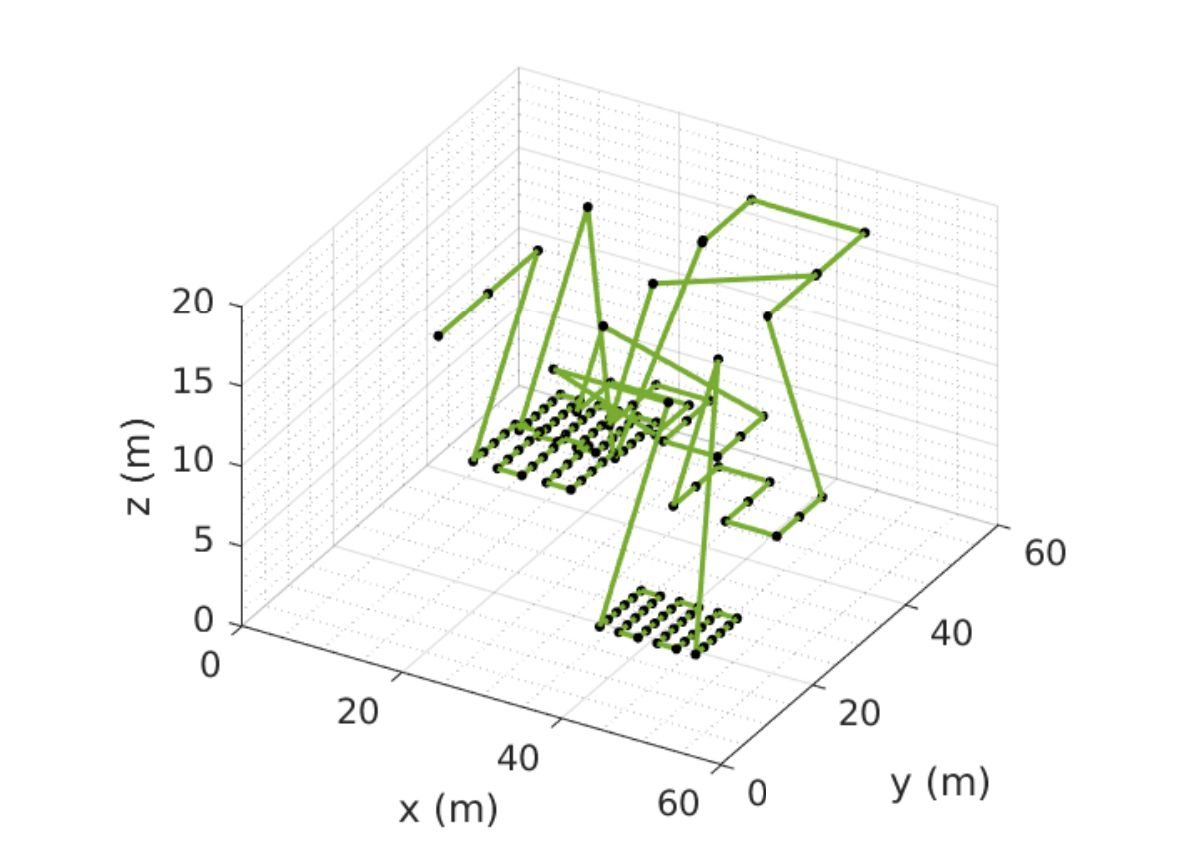}
   \includegraphics[width=0.3\linewidth]{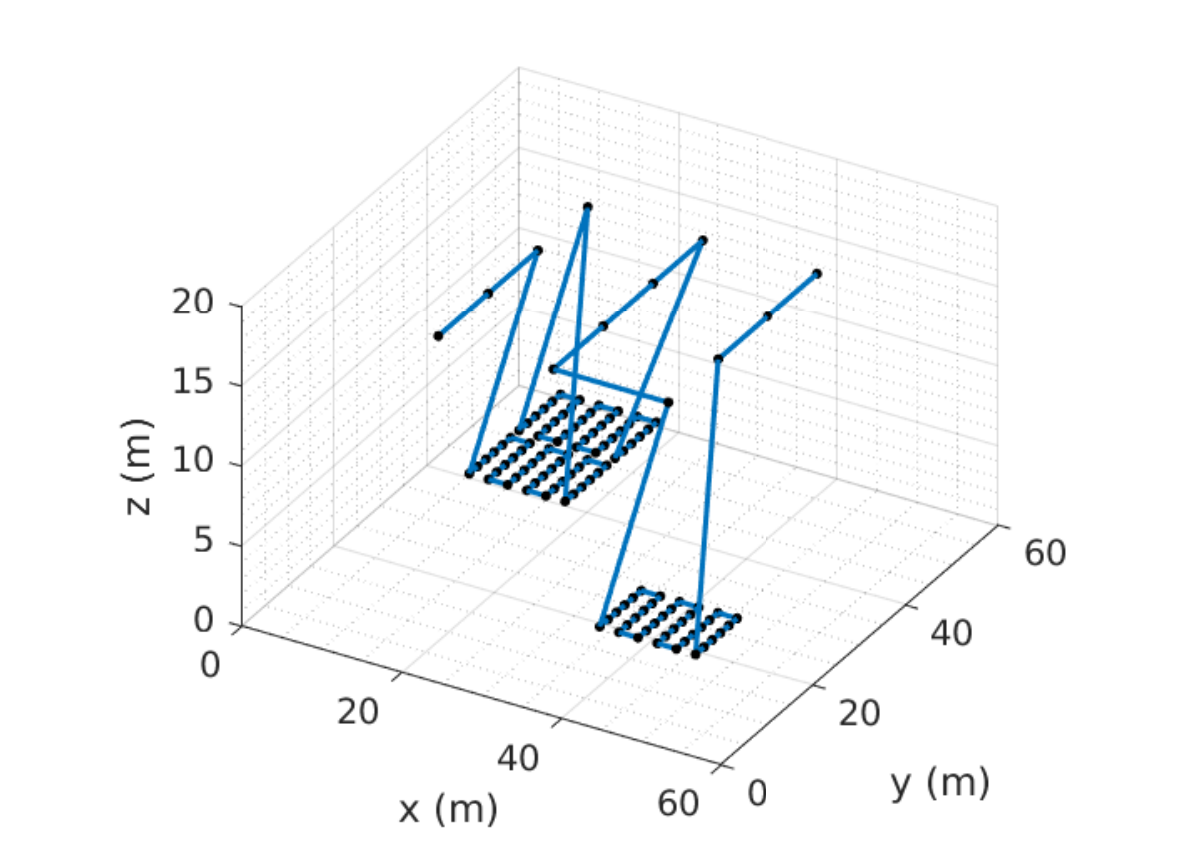}
  \caption{Visual comparison of trajectories traveled by the UAV over a field using different planning strategies. The coverage paths (left) are restricted to fixed heights and cannot map targeted areas of interest. The linear decision function (middle) enables adaptive planning, but it is continuous with respect to altitude and leads to sudden jumps. Our adaptive approach overcomes this issue, leaving the path less often and more purposefully at selected heights for more efficient mapping. The black spheres indicate measurement points.}
  \label{fig:waypoints}
\end{figure*}

\section{Experimental Results}
\label{sec:experimental}
We validate our proposed algorithm for online adaptive path planning on the application of UAV-based crop/weed semantic segmentation. The goal of our experiments is to demonstrate the benefits of using our adaptive strategy to maximize segmentation accuracy in missions while keeping a low execution time. Specifically, we show results to support two key claims: our online adaptive algorithm can (i) map high-interest regions with higher accuracy and (ii) improve segmentation accuracy while keeping a low execution time with respect to the baselines described in \secref{subsec:baselines}.

\subsection{Dataset}

To evaluate our approach, we use the WeedMap dataset~\cite{Sa2018b}. It consists of 8 different fields collected with two different having different channels, it also provides pixel-wise semantic segmentation labels for each of the 8 fields. In this study, we focus only on the 5 fields having RGB information. We split the 5 fields into training and testing sets (\figref{fig:dataset_split}). One of the training fields is used to initialize the decision function that shapes altitude selection in the adaptive strategy, as described in \secref{subsec:online}. For each experiment in the following sub-sections, we test our approach and the baselines, see \secref{subsec:baselines}, on each field once, and then report the average among each run.

\subsection{Baselines}
\label{subsec:baselines} 
To evaluate our proposed approach, we compare it against two main baselines. The first one is the standard lawn-mower strategy where a UAV covers the entire field at the same altitude, for this strategy we use consider five different altitudes resulting in GSD $\in \{ 1.0, 1.5, 2.0, 2.5, 3.0 \} \frac{\text{cm}}{\text{px}}$. The lawnmower strategy with a fixed GSD of $3.0 \frac{\text{cm}}{\text{px}}$ corresponds to the initial plan for our strategy described in \secref{subsec:planning}.
The second baseline is defined by \textit{only} initializing the UAV behavior as described in \secref{subsec:init} and without adapting the strategy online using the decision function as new segmentations arrive. We refer to this strategy as ``Non Adaptive''. This benchmark allows us to study the benefit of adaptivity obtained by using our proposed approach (``Adaptive'').

\subsection{Metrics}
\label{subsec:metrics}

Our evaluation consists of two main criteria: segmentation accuracy and mission execution time. For execution time, we compute the total time taken by the UAV to survey the whole field, including the time needed to move between waypoints, segment a new image, and plan the next path. To assess the quality of the semantic segmentation we use the mIoU metric defined in~\eqref{eq:iou}.

\subsection{Field Segmentation Accuracy vs Execution Time}
\label{sec:semseg_acc}
The first experiment is designed to show that our proposed strategy obtains higher accuracy while keeping low execution time. We show such results in \figref{fig:results}. For each strategy, we compute the mIoU (over the entire field) and the execution time needed by the UAV to complete its path. The adaptive strategy crosses the line defined by the lawnmower strategies at different altitudes, meaning that it can achieve better segmentation accuracy while keeping a lower execution time. The non-adaptive strategy instead lies under the curve, failing to overtake the lawn-mower strategy. We plot exemplary paths results from the different strategies in \figref{fig:waypoints}; on the left, we show the lawn-mower strategy with altitudes corresponding to GSDs of $1.0\frac{\text{cm}}{\text{px}}$ and $3.0\frac{\text{cm}}{\text{px}}$, while the middle and right plots show the paths resulting from non-adaptive and adaptive strategy, respectively. 

\subsection{Per-Image Segmentation Accuracy vs Altitude}
\label{sec:semseg_acc}
\begin{figure}[t] 
  \centering
   \includegraphics[width=0.9\linewidth]{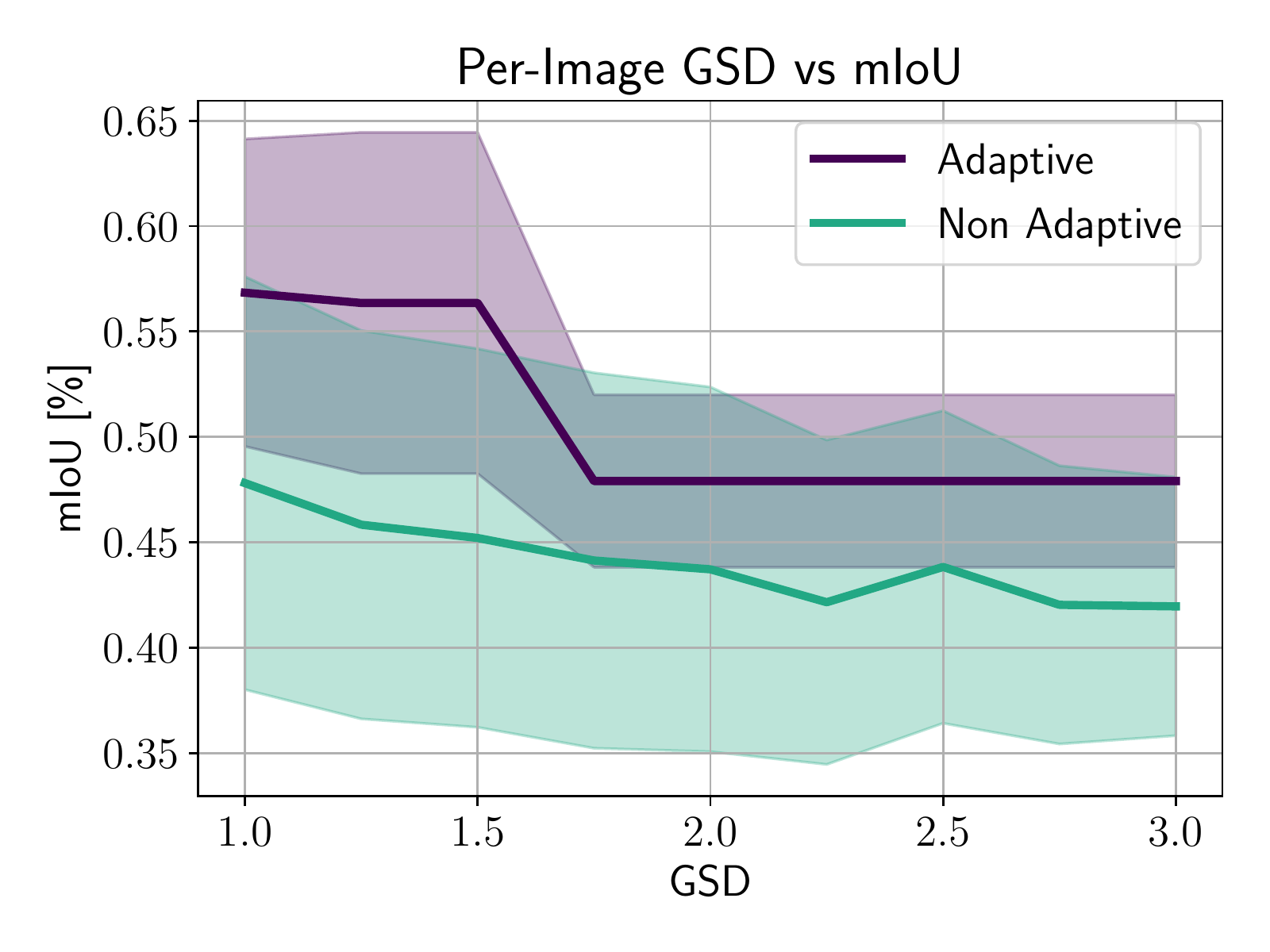}
  \caption{Mean and standard deviation of the per-image statistics for semantic segmentation. Our adaptive strategy leads to better performance when scouting the field at low altitudes.}
  \label{fig:acc_vs_gsd}
\end{figure}

\begin{figure}[t]
  \centering
   \includegraphics[width=0.9\linewidth]{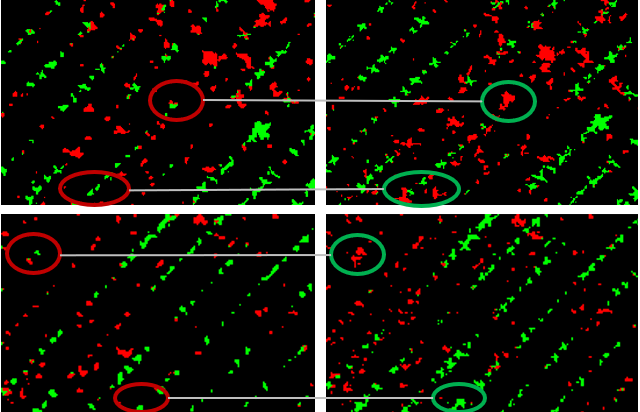}
  \caption{Qualitative field segmentation results using the non-adaptive strategy (left) and the proposed adaptive strategy using our decision function (right) for path planning. The circled details demonstrate that our adaptive planning approach enables targeted high-resolution segmentation to better capture finer plant details.}
  \label{fig:results2}
\end{figure}

The second experiment shows the ability of our approach to achieve targeted semantic segmentation when compared to the non-adaptive strategy. At this stage, we compute mIoU for each image that contributes to the final segmentation of the whole field. This will give us a way to evaluate the efficiency of our adaptation strategy. We then visualize the mean and standard deviation. As can be seen in \figref{fig:acc_vs_gsd}, our adaptive strategy provides higher per-image accuracies when the UAV is scouting the field at low altitudes. This entails that, with our strategy, the UAV invests time resources in a more proficuous manner. We show a qualitative comparison of the per-image semantic masks in \figref{fig:results2}.

\section{Conclusion}
\label{sec:conclusion}

In this paper, we presented a new approach for efficient multi-resolution mapping using UAVs for semantic segmentation. We exploit prior knowledge and the new incoming segmentations in a way, that we get a decision function with a shape that produces a flight path, leading to a performance gain in terms of segmentation accuracy while keeping comparatively short execution time. The resulting map is mapped with different resolutions, depending on the information content of a corresponding area. We believe that our approach opens a direction for efficient UAV mapping purposes, especially in precision agriculture. Further investigations on less homogeneous field structures are to be carried out to refine the approach.

\bibliographystyle{plain_abbrv}

\bibliography{new}

\end{document}